\documentclass[runningheads]{llncs}

\usepackage{eccv}

\usepackage{eccvabbrv}

\usepackage{graphicx}
\usepackage{booktabs}
\usepackage{xcolor}
\usepackage{wrapfig}  
\usepackage{tabularx}
\usepackage{algorithm}
\usepackage{algorithmic}

\usepackage[accsupp]{axessibility}  

\usepackage{hyperref}

\usepackage{orcidlink}

\begin{document}

\title{AnyStyle: A Single LoRA is Sufficient for Image-Guided Style Transfer} 

\titlerunning{AnyStyle}

\author{Yongwen Lai \orcidlink{0009-0007-2597-5065} \and
Chaoqun Wang\thanks{Corresponding author.} \orcidlink{0000-0002-4649-5518}}

\authorrunning{Yongwen Lai, Chaoqun Wang}


\institute{School of Artificial Intelligence, South China Normal University, Guangzhou, China 
}

\maketitle


\begin{figure*}[h]
    \centering
    \vspace{-20pt}
    \includegraphics[width=0.85\textwidth]{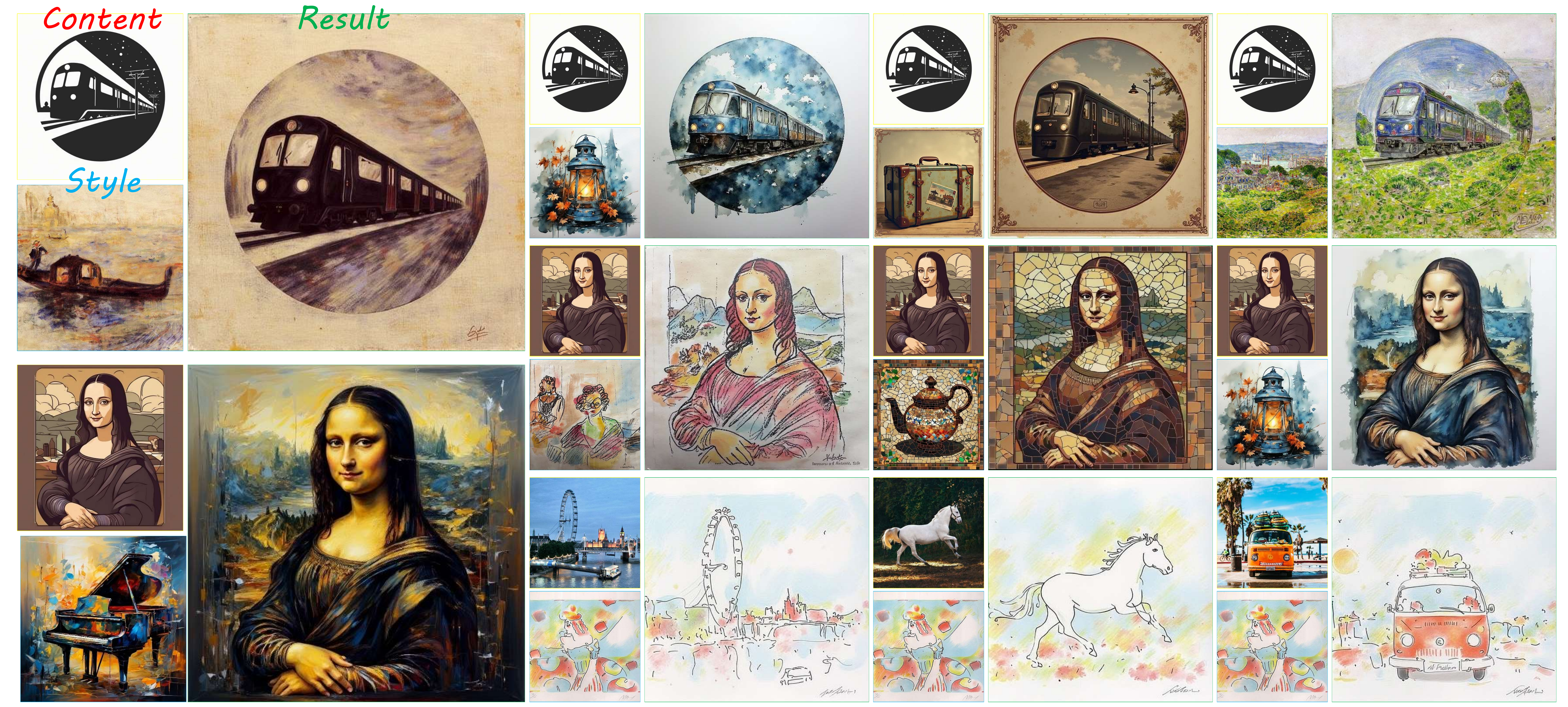}
    \caption{\textbf{Style transfer results of our method.} In each set of images, the top-left image denotes the content reference, the bottom-left image denotes the style reference, and the image on the right presents the generated results. The results demonstrate that our method consistently achieves both content preservation and style alignment across diverse inputs, including real-world photographs and painted images.}
    \label{fig:teaser}
    \vspace{-30pt}
\end{figure*}

\begin{abstract}
Image-guided style transfer aims to apply the artistic characteristics of a style image to a content image while preserving its semantic structure and layout. Despite advances in diffusion-based methods, existing approaches often face challenges in disentangling content and style, particularly when independently optimized adapters are naively combined, causing conflicts between adapters and limiting controllability over the content-style balance in inference. We further demonstrate that training-free structural guidance directly derived from the content image through the internal attention of pre-trained model outperforms a dedicated content LoRA adapter in terms of structural fidelity and computational efficiency. Building on these observations, we propose AnyStyle, a streamlined framework for image-guided style transfer. The framework adopts a unified single-adapter paradigm for coherent style capture from the style image and incorporates training-free structural guidance from the content image, thus avoiding complex entanglement between multiple adapters and improving controllability and stability. Extensive experiments show that our method delivers competitive quantitative performance and significantly improved perceptual quality. Code is available at \url{https://github.com/Yvan1001/AnyStyle}.
\keywords{Style Transfer \and Low-Rank Adaptation \and Rectified Flow}
\end{abstract}

\section{Introduction}

\begin{wrapfigure}{r}{0.45\textwidth}  
    \vspace{-22pt}   
    \centering
    \includegraphics[width=\linewidth]{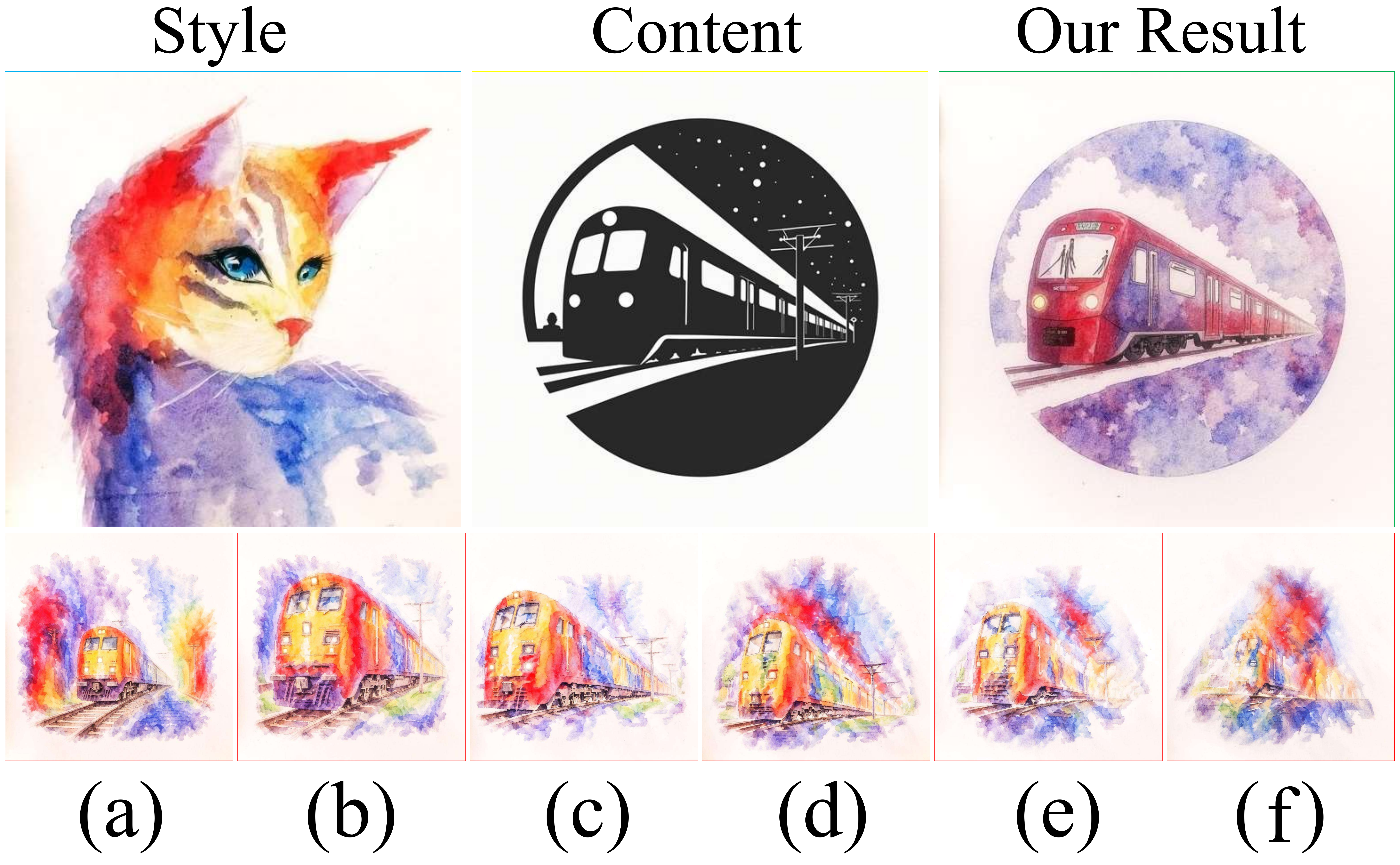} 
    \caption{The first row shows the input and the output of our method. Columns (a–f) present the results of the dual-LoRA approach under different fusion coefficients during inference(Adjust the coefficients to shift from style dominance to content dominance). As illustrated, the results remain unsatisfactory across different coefficient settings, achieving a balanced trade-off with the dual-LoRA method is difficult.}
    \label{fig:issues_pictures}
    \vspace{-10pt}   
\end{wrapfigure}

Image-guided style transfer\cite{frenkel2024implicit, jia2024dginstyle, nikolaidou2024diffusionpen, cui2024instastyle, li2024styletokenizer} aims to apply the artistic characteristics of a style reference image to a content reference image while preserving as faithfully as possible the semantic structure and layout of the content reference image. This task holds significant value and has broad applications in digital art creation, image editing, and cross-modal content generation. With the evolution of deep learning, early methods\cite{gatys2016styletransfercnn,jing2020neural,lai2017deeplaplacian,li2017universalstyle} based on convolutional neural networks have achieved pioneering results by iteratively minimizing a joint content-style loss. However, these approaches exhibit strong dependency on data distributions specific to particular styles, and collecting high-quality datasets for style transfer remains challenging in practice. Subsequent advances in generative adversarial networks and diffusion models substantially improve generation quality and generalization. Modern approaches built upon large-scale pre-trained text-to-image diffusion models\cite{flux2024,esser2024scaling} now represent the dominant paradigm for style transfer. 

In recent years, the fine-tuning technique Low-Rank Adaptation (LoRA) has been widely adopted for personalized style generation. Taking advantage of this convenience, a prominent line of work\cite{frenkel2024implicit,liu2025unziplora,chen2025consislora,yang2025splitflux} employs a dual-LoRA strategy for the style transfer task. In this approach, separate LoRA weights are independently fine-tuned for content and style and then combined via simple weight fusion to denoise in inference. Although such methods advance the field, our empirical analysis reveals several inherent limitations. 1) Independently optimized adapters often cause complex entanglement between content and style adapters. Their fusion requires delicate tuning of mixing coefficients, rendering the content-style trade-off difficult, as illustrated in \cref{fig:issues_pictures}. Meanwhile, although modern diffusion models are often described as exhibiting a layered processing pattern, in which earlier layers encode low-level structural information and later layers emphasize higher-level semantics and style, such a division is only partially valid in practice. Feature representations of content and style in diffusion models remain inherently entangled across layers, which challenges the assumption that content and style can be reliably disentangled through a dual-LoRA design. 2) A dedicated content LoRA typically underperforms direct utilization of training-free structural guidance derived from the content reference image through the internal attention of the pre-trained model in structural fidelity and perceptual quality, while adding fine-tuning overhead. Under these circumstances, training-free structural guidance thus provides a more direct, efficient, and effective means of content preservation in the style transfer task.

Building upon these findings, we propose AnyStyle, an image-guided style transfer framework that combines a single-style LoRA with an attention modulation mechanism. The complete framework consists of two stages: During fine-tuning, we adopt a single-adapter design in place of the dual-LoRA paradigm, fine-tuning only one LoRA targeted at layers more suitable for encoding style, thereby effectively enhancing the ability to generate consistent style. During inference, training-free structural guidance is extracted from the content reference image and dynamically incorporated to reliably preserve content layout and structure. This design effectively avoids the complex entanglement and uncertainty introduced by multi-adapter fusion and achieves improved coordination among textual-prompt adherence, content fidelity, and style consistency. The main contributions of our work are as follows:
\begin{itemize}
\item Through experimental analysis, we uncover key limitations of the dual-LoRA strategy in style transfer and validate the superiority of training-free structural guidance from content reference image for content preservation.
\item We introduce a simple yet effective pipeline that integrates a single LoRA with attention modulation and enhances the controllability and stability of style transfer. 
\item We demonstrate competitive performance against existing methods in both quantitative metrics, visual results, and user study evaluations.
\end{itemize}

\section{Related Work}
\subsection{Style Transfer}

Image-guided style transfer\cite{efros2001image,hertzmann2001image} is a classic task, which aims to apply the artistic style of a style reference image to a content reference image while preserving the semantic structure of the object and layout in the content reference image as faithfully as possible. The early approaches are primarily based on convolutional neural networks\cite{gatys2016styletransfercnn,jing2020neural,lai2017deeplaplacian,li2017universalstyle,li2018closedsolution,lu2019universal,johnson2016perceptual}, achieving a content-style balance through iterative optimization of a combined loss function. These methods mark a significant breakthrough in the style transfer task, but suffer from strong dependency on datasets and limit output diversity. With the advent of deep generative models, many methods\cite{chong2022jojogan, gal2022stylegan, ojha2021few} based on generative adversarial networks have emerged. These approaches substantially improve generation quality and generalization. However, they typically require large domain-specific datasets, incur high retraining costs, and are prone to mode collapse or structural distortion in semantically complex scenes. 

In recent years, diffusion models\cite{ho2020ddpm, Rombach2022highldm, song2020denoising} have fundamentally transformed the paradigm of style transfer by leveraging the powerful generative priors of pre-trained text-to-image models such as Stable Diffusion 1.5\cite{Rombach2022highldm}, SDXL\cite{podell2024sdxl}. In more recent advances, diffusion model backbones have transitioned from traditional U-Net\cite{ronneberger2015u} to more powerful Transformer\cite{vaswani2017attention}.  Models including DiT\cite{peebles2023dit}, SD3\cite{esser2024scaling}, and FLUX\cite{flux2024} have enabled diffusion models to scale toward high-resolution and semantically complex image generation. By exploiting the rich prior knowledge embedded in pre-trained diffusion models, recent methods investigate image stylization and editing by modifying the generation process through fine-tuning and training-free methods. For example, InST\cite{zhang2023InST} proposes a textual inversion-based method that encodes a target style into learned textual embeddings. StyleDiffusion\cite{wang2023stylediffusion} seeks to separate style from content by incorporating a CLIP-based\cite{radford2021learning} style disentanglement loss during the fine-tuning of diffusion models for style transfer. Style-Friendly\cite{choi2024style} aggressively shifts the signal-to-noise ratio distribution toward higher noise levels during fine-tuning to focus on noise levels where stylistic features emerge. During the same period, many training-free methods\cite{wang2024instantstyle,hertz2024style,aravanis2025onlystyle} emerge. Specifically, Only-Style\cite{aravanis2025onlystyle} and StyleAligned\cite{hertz2024style} achieve style consistency across a sequence of generated images by combining attention feature sharing with the AdaIN\cite{huang2017adain} mechanism. However, these methods do not explicitly incorporate content control, which may lead to the loss of structural information from the content reference image.

\subsection{Attention and Structural Guidance}
In image generation tasks, guiding the sampling process with attention and structural information is a widely adopted strategy. Methods such as Ctrl-X\cite{lin2024ctrlx} achieve structural control via a dedicated module but require clean structural input, imposing restrictive assumptions in practice. FreeControl\cite{lin2025freecontrol} performs one-step attention extraction from a single, optimally chosen key timestep and reuses it throughout denoising to enable efficient structural guidance. P2P~\cite{hertz2023prompttoprompt} injects cross-attention maps from the reconstruction branch into the editing branch. RF-Solver-Edit~\cite{wang2024taming} and PnP~\cite{tumanyan2023plug} reuse values from self-attention layers of the source inversion. Training-based methods such as ControlNet\cite{zhang2023adding} and T2I-Adapter\cite{mou2024t2i} leverage condition maps to guide spatial structure, achieving strong low-level alignment but requiring retraining per condition and per model, which incurs substantial computational cost and limits robustness to noisy inputs. Higher-level approaches, including GLIGEN\cite{li2023gligen} and IP-Adapter\cite{ye2023ip}, enable layout-aware or visual conditioning, yet still depend on specific training and provide limited fine-grained structural control. In addition, a growing number of studies\cite{cho2024preservation,chung2024styleid,wang2024instantstyle-plus,xu2024freetuner,zhang2024artbank} have investigated methods to improve content preservation by enforcing structural and semantic consistency during generation. Previous methods tend to emphasize either content preservation or style consistency, making it challenging to achieve a balanced coordination of both.

\subsection{LoRA-based Image Style Transfer}
Early work such as DreamBooth\cite{ruiz2023dreambooth} initiate this line of research by fine-tuning diffusion models on a small set of subject-specific images to encode identity information, but at the cost of substantial computational overhead and increased risk of overfitting. To address these limitations, LoRA\cite{hu2022lora} enables parameter-efficient fine-tuning by applying low-rank updates to selected model layers, significantly reducing memory and computation while maintaining generation quality. Consequently, LoRA-based methods\cite{shah2024ziplora,frenkel2024implicit, liu2025unziplora,chen2025consislora, yang2025splitflux} have emerged as the predominant approach for style transfer, supporting a wide range of applications. Specifically, ZipLoRA\cite{shah2024ziplora} enables flexible subject–style composition by fusing independently trained LoRA weights (one for
content, one for style) through column-wise merging coefficients, but each fusion still requires additional optimization, limiting reuse. B-LoRA\cite{frenkel2024implicit} leverages block specialization in the SDXL to achieve implicit content–style disentanglement via joint optimization on selected blocks, although coarse block partitioning often leads to subject leakage. UnZipLoRA\cite{liu2025unziplora} further improves disentanglement by introducing prompt, column, and fine-grained block separation.
Despite these advances, challenges remain in generating stylized images that preserve content structure and align with the desired style. 

In our work, we adopt only one LoRA targeted at layers more suitable for encoding style as the foundation for achieving style consistency. Building upon this, we guide the sampling process using attention modulation that incorporates training-free structural guidance from a content reference image, thereby enabling collaborative stylization that jointly preserves content and style.

\section{Preliminaries}
\subsection{Rectified Flow Model}
Rectified flow\cite{liu2022flow} is a continuous generative framework that facilitates smooth and efficient transport between the image data distribution $X$ and a Gaussian distribution $N \sim \mathcal{N}(0, I)$. Unlike traditional diffusion models that rely on stochastic multi-step transitions between noise and data, rectified flow establishes a deterministic and nearly linear trajectory for mapping noise to data, which is governed by an ordinary differential equation (ODE) parameterized by a neural network $\theta$. The intermediate latent state at continuous timestep $t \in [0, 1]$ is formulated as a linear interpolation: 
\begin{equation}
X_t = (1-t)\cdot X + t\cdot N,
\label{eq:linear interpolation}
\end{equation}
where $t=1$ corresponds to pure noise and $t=0$ denotes the clean data domain. During inference, starting from a Gaussian noise sample, the pre-trained model iteratively predicts the velocity $v_\theta$ at each discrete timestep and updates the latent state accordingly. Specifically, the update of the latent state $Z_t$ follows the recurrence relation:
\begin{equation}
Z_{t_{i-1}} = Z_{t_i} + (t_{i-1} - t_i) v_\theta(Z_{t_i}, t_i).
\end{equation}
After completing the iterative denoising process, the final latent state is decoded into the pixel space through a decoder to obtain the generated image.

\subsection{FlUX Architecture and LoRA Fine-tuning}

\begin{wrapfigure}{r}{0.45\textwidth}  
    \centering
    \includegraphics[width=\linewidth]{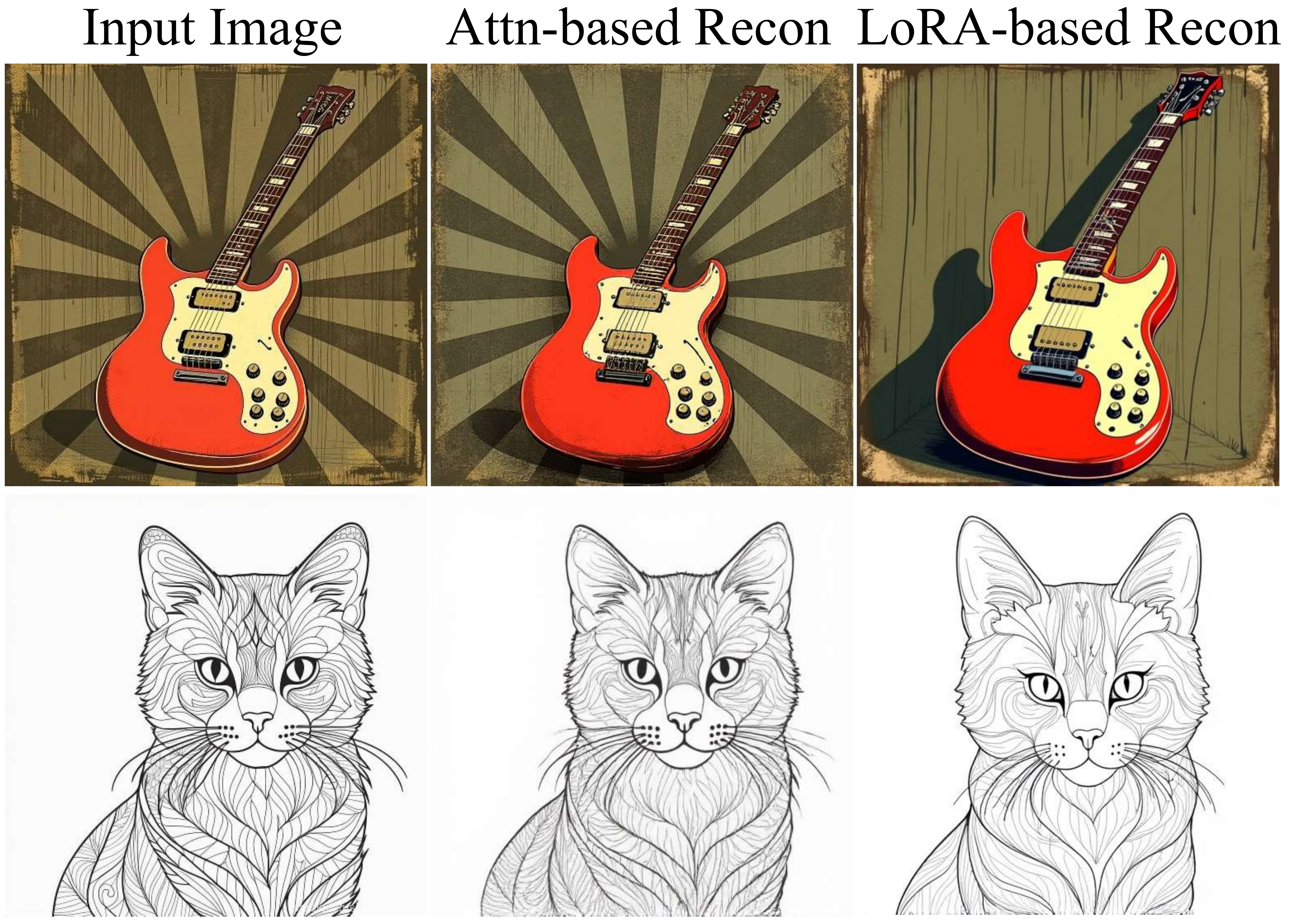} 
    \caption{Reconstruction. Attention-based reconstruction preserves structure and perceptual quality with a single inference, while LoRA-based requires extra fine-tuning and yields lower fidelity.}
    \label{fig:motivation}
\end{wrapfigure}


\textbf{FLUX Architecture}. FLUX\cite{flux2024} is a state-of-the-art text-conditional generative framework built upon the rectified flow paradigm and the DiT\cite{peebles2023dit} backbone. Its core architecture comprises 19 double-stream transformer blocks and 38 single-stream transformer blocks, which collaboratively process multimodal inputs. The input text prompt is encoded via CLIP\cite{radford2021learning} and T5\cite{raffel2020exploring}, ensuring that semantic guidance is effectively embedded into the generation process.

\noindent
\textbf{LoRA Fine-tuning}. To efficiently adapt such large-scale pre-trained models to task-specific scenarios like style transfer, LoRA\cite{hu2022lora} is a parameter-efficient fine-tuning strategy that avoids full model retraining. For a pre-trained linear transformation layer, its weight matrix is denoted as \( W_0 \in \mathbb{R}^{d_{\text{out}} \times d_{\text{in}}} \) where \( d_{\text{in}} \) and \( d_{\text{out}} \) denote the dimensions of the input and output of the layer. This matrix maps an input vector \( x \in \mathbb{R}^{d_{\text{in}}} \) to an output vector \( y \in \mathbb{R}^{d_{\text{out}}} \) following the linear transformation \( y = W_0 x \). LoRA introduces a low-rank residual update to capture task-specific adjustments, with the effective fine-tuned weight matrix and its low-rank decomposition jointly expressed as:
\begin{equation}
W = W_0 + \Delta W,\quad \Delta W = BA,
\end{equation}
where \( W_0 \) is the original frozen pre-trained weight matrix, \( \Delta W \) is the low-rank update introduced by LoRA fine-tuning. The rank of the low-rank decomposition \( r \) is defined as a small value typically satisfying \( r \ll \min(d_{\text{in}}, d_{\text{out}}) \) to minimize the number of trainable parameters. The matrices \( A \in \mathbb{R}^{r \times d_{\text{in}}} \) and \( B \in \mathbb{R}^{d_{\text{out}} \times r} \) form the low-rank factorization of \( \Delta W \), and \( W \) denotes the effective weight matrix used during inference. During training, only the low-rank matrices \( A \) and \( B \) are updated, while the original pre-trained weights \( W_0 \) remain frozen. This drastically reduces the number of trainable parameters and mitigates the risk of overfitting. During inference, the low-rank updates can be merged into the original weight matrices without incurring additional computational overhead, ensuring efficient deployment.

\begin{figure*}[t]
    \centering
    \includegraphics[width=\textwidth]{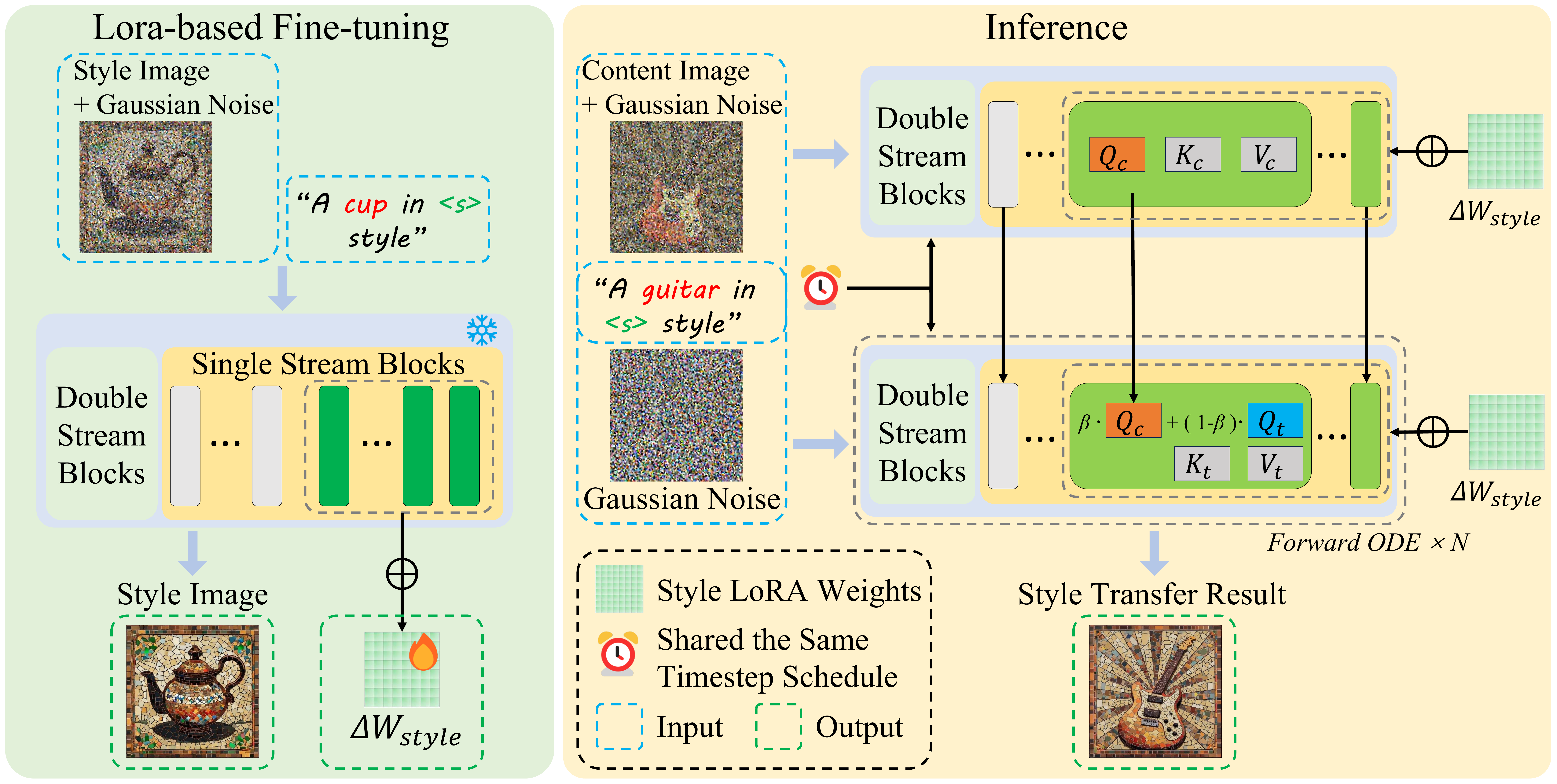}
    \caption{\textbf{Overview of the AnyStyle framework.} Given a content reference image and a style reference image, together with a prompt such as "A <c> in <s> style", AnyStyle produces stylized results that preserve the structural integrity of the original object while ensuring stylistic consistency. The left panel illustrates the fine-tuning stage, whereas the right panel presents the inference stage. During inference, the style LoRA weights are integrated into the corresponding DiT blocks to enforce consistent style generation. In addition, an attention modulation module is employed to guide structural features, enabling effective style transfer while maintaining content fidelity.}
    \label{fig:pipeline}
\end{figure*}

\section{Method}
Existing style transfer methods typically adopt a dual-LoRA paradigm, separating content and style adaptation across different layer ranges. However, such designs rely on imperfect layer-wise disentanglement and often require delicate weight balancing due to the complex coupling relationship between content and style. To address these limitations, we propose a streamlined single-LoRA framework with attention modulation termed AnyStyle for image-guided style transfer. 

\subsection{Analysis of Dual-LoRA for Style Transfer}
FLUX\cite{flux2024}, based on a DiT\cite{peebles2023dit} architecture, features double-stream and single-stream transformer blocks in its core denoising component. Through extensive experimentation and analysis, we observe that early single-stream transformer blocks primarily capture low-level structural details, whereas later blocks tend to encode higher-level semantic and stylistic attributes. Consistent with the above findings, many prior works\cite{frenkel2024implicit,liu2025unziplora,chen2025consislora,yang2025splitflux} adopt dual-LoRA architectures for style transfer by fine-tuning separate LoRA adapters\cite{hu2022lora} for content and style in different layer ranges, and subsequently fuse the learned weights.

To evaluate the effectiveness of the content LoRA in the dual-LoRA strategy for the style transfer task, we analyze its assumptions and limitations by comparing two content reconstruction strategies. The first strategy fine-tunes content LoRA weights on early single-stream blocks (1–10) using a single reference image and applies them for reconstruction from noise. The second adopts an attention-based approach that adds noise to the content reference image, records query, key, and value tensors during denoising, and uses them to guide reconstruction from random noise without any LoRA weights. The attention-based method achieves superior structural fidelity and perceptual quality while eliminating additional fine-tuning overhead. The corresponding quantitative results are reported in \cref{tab:verify_single_lora}. Detailed visual comparisons of the reconstruction can be found in \cref{fig:motivation}. From the visualization, the attention-based approach demonstrates significantly superior reconstruction quality while requiring only a single inference process. In contrast, the LoRA-based approach not only produces lower reconstruction fidelity but also incurs additional time for LoRA fine-tuning. These results suggest that content LoRA weights provide limited value for content preservation. Recording query, key, and value tensors during denoising captures more precise structural information than fine-tuning a content LoRA module, making content LoRA largely redundant. Furthermore, dual-LoRA architectures suffer from complex entanglement: independently optimized content and style adapters remain coupled in single-stream blocks, and their fusion requires careful coefficient tuning, often leading to suboptimal trade-offs.

\begin{table}[t]
\centering
\footnotesize
\caption{Quantitative comparison of image reconstruction under two strategies. In this context, "Attn-based" refers to the approach that guides image reconstruction using the recorded query, key, or value tensors, whereas "LoRA-based" refers to the approach that guides image reconstruction using the fine-tuned LoRA weights.}
\label{tab:verify_single_lora}
\begin{tabularx}{\linewidth}{lXXXXX}
\toprule
Reconstruction\quad\quad
& PSNR($\uparrow$)
& DINO-Content($\uparrow$)
& DS-Content($\downarrow$)
& MSE($\downarrow$)
& LPIPS($\downarrow$) \\
\midrule
Attn-based 
& 18.0029 \textcolor{green}{$\uparrow$}
& 0.9743 \textcolor{green}{$\uparrow$}
& 0.0617 \textcolor{green}{$\downarrow$}
& 0.0754 \textcolor{green}{$\downarrow$}
& 0.2438 \textcolor{green}{$\downarrow$} \\
LoRA-based 
& 11.7899
& 0.9567
& 0.1273
& 0.3137
& 0.4553 \\
\bottomrule
\end{tabularx}
\end{table}


\subsection{AnyStyle for Style Transfer}
Motivated by these observations, we propose AnyStyle, a streamlined and effective framework for style transfer that employs only a single layer-specific LoRA for style alignment combined with a complementary attention modulation for content preservation. This design can avoid the complexities and instability of dual-LoRA fusion while achieving robust and predictable control over style and content. The overall pipeline is illustrated in \cref{fig:pipeline}.

For style alignment, we apply LoRA exclusively to the later single-stream transformer blocks (11–38) in FLUX, targeting the projection matrices of self-attention layers (query, key, value, and output projections), where corresponding low-rank updates are optimized to improve the ability to generate specific styles. Specifically, by differentiating the interpolation formula in \cref{eq:linear interpolation} with respect to timestep $t$, the underlying flow ODE can be derived as:  
\begin{equation}
\frac{d X_t}{dt} = X - N,
\end{equation}
which defines the target velocity field that the neural network with style LoRA weights is designed to approximate, and the objective of fine-tuning is to minimize the mean squared error between the predicted velocity $v_{\theta,\Delta W_{style}}(X_t, t)$ and the target velocity across the entire time interval, expressed as:
\begin{equation}
\min_\theta \int_0^1 \mathbb{E}\left[ \left\| (X - N) - v_{\theta,\Delta W_{style}}(X_t, t) \right\|_2^2 \right] dt.
\end{equation}
Each LoRA module is fine-tuned independently following the DreamBooth\cite{ruiz2023dreambooth} objective, which reconstructs the reference image from Gaussian noise conditioned on a textual prompt. This single, targeted LoRA efficiently infuses stylistic priors, including colors, textures, and artistic patterns, into the layers where style representations naturally dominate. In inference, the fine-tuned style LoRA weights are loaded into the AnyStyle pipeline and integrated with standard sampling. By replacing the dual-LoRA strategy with this single-LoRA plus attention modulation framework, we mitigate complex entanglement and unstable trade-offs between content and style arising from the simple fusion of independently optimized adapters, supporting flexible stylization of arbitrary content-style pairs and allowing the target result to consistently maintain the intended structure while effectively applying the desired stylistic characteristics.

\begin{algorithm}[t]
\centering
\footnotesize
\raggedright
\noindent\textbf{Input:} $Z^{c}, Z^{s}, \{t_i\}_{i=0}^T, n,\beta_{max}, \beta_{min}$ \\
\noindent\textbf{Output:} Style Transfer Result $Z_0$ \\
$\Delta W \leftarrow \text{DreamBooth}(Z^{s})$;\;$\Theta \leftarrow \theta + \Delta W$\; \\
Initialize $Z_T \sim \mathcal{N}(0,I)$; $ \epsilon \sim \mathcal{N}(0,I)$ \\
\textbf{for} $i=n$ to $1$: \\
\quad $\beta \leftarrow \beta_{max} \cdot ({t_i}/T) + \beta_{min} \cdot (1 - {t_i}/T)$ \\
\quad $Z^{c}_{t_i} \leftarrow (1-{t_i}) \cdot Z^{c} + {t_i} \cdot \epsilon$ \\
\quad $V^{c}_{t_i} \leftarrow V_{\Theta}(Z^{c}_{t_i},{t_i}) \implies \text{extract } Q_{\text{c}}$; \quad $V_{t_i} \leftarrow V_{\Theta}(Z_{t_i},{t_i},Q_{\text{c}})$ \\
\quad $Z_{t_{i-1}} \leftarrow Z_{t_i} - \Delta {t} \cdot V_{t_i}$ \\
\textbf{return} $Z_0$ \\
\caption{Full AnyStyle Algorithm}
\label{alg:anystyle}
\end{algorithm}

\subsection{Structural Guidance with Attention Modulation}
For content preservation, methods that rely solely on fine-tuning an additional content adapter under the dual-LoRA paradigm frequently result in distorted layouts or loss of object structure when applying aggressive style changes.

\begin{figure*}[t]
    \centering
    \includegraphics[width=0.8\textwidth]{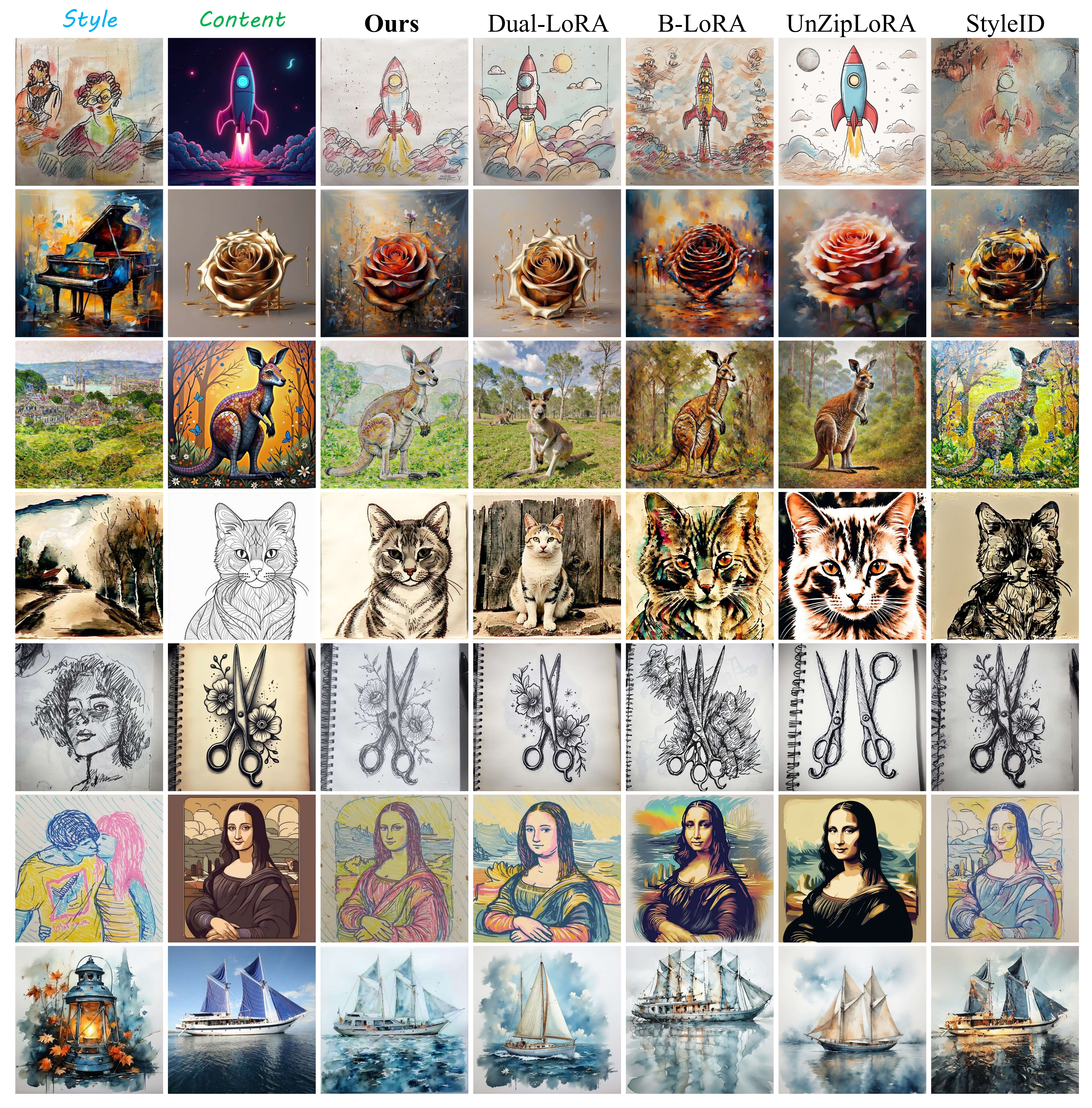}
    \caption{Qualitative comparisons. The first and second columns correspond to the style reference image and the content reference image, respectively. The subsequent columns present the visual results obtained using different style transfer methods.}
    \label{fig:sota_visualization}
    \vspace{-15pt}   
\end{figure*}

To overcome this limitation, we design an attention modulation that directly extracts training-free structural guidance from the content image itself, avoiding separate training or external adapters. The key is to leverage the inherent attention dynamics of the content image under controlled noise. Specifically, we introduce an auxiliary denoising path: at each timestep $  t  $, the input latent of the DiT model $Z_t^c$ is $ (1-t)\cdot Z^c + t\cdot N$, where $Z^c$ represents the latent of the content image. During the denoising process, attention tensors from self-attention layers are recorded and indexed by timestep and block, and then selectively mixed into the main generation path. For each attention call at the corresponding timestep and block, the content query $Q_c$ is retrieved and blended with the current query $Q_t$ according to:
\begin{equation}
Q_t = \beta \cdot Q_c + (1 - \beta) \cdot Q_t,
\end{equation}
where $\beta$ controls the strength of the structural guidance. Key and value tensors remain unchanged to prevent any stylistic leakage from the content reference image. Ablation studies confirm that modulating only the query tensors is sufficient for effective content preservation, as queries primarily encode "what to attend to", such as layouts and objects, whereas modifying key or value tensors tends to introduce unwanted stylistic elements from the content reference image. Through thorough experimentation, this selective approach provides precise structural guidance without compromising style alignment. To ensure adaptive guidance across the denoising process, we apply time-dependent weighting:
\begin{equation}
\beta = \beta_{\max} \cdot (t/T) + \beta_{\min} \cdot (1 - t/T),
\end{equation}
where $T$ denotes the first number in denoising steps, $\beta_{\max}$ and $\beta_{\min}$ denote the maximum and minimum modulation strengths applied at the beginning and end of the denoising process, respectively. Actually, systematically decreasing $\beta$ values from $1$ to $0$ results in a smooth and continuous transition from heavily content-preserving visual generations to style-dominant outputs. For details, please refer to the supplementary materials. This design delivers stronger modulation in early timesteps, where noise is dominant and the content structure is most vulnerable, and weaker modulation in later timesteps, which allows stylistic refinement once the core content structure has been established. The modulation is applied across all 38 single-stream transformer blocks to achieve holistic and consistent content preservation. These timestep-dependent attention signals then dynamically guide the main generation path, preserving content structure while fully respecting the style imposed by the style LoRA adapter.
Algorithm~\ref{alg:anystyle} summarizes the proposed Anystyle. Detailed algorithmic procedures and variable definitions are provided in the supplementary materials.
\begin{table}[t]
\centering
\footnotesize
\caption{Quantitative comparison across different style transfer methods. The style transfer results are assessed separately against the content reference image and the style reference image, yielding corresponding content and style scores. In addition, a text consistency score is computed to evaluate whether the generated images are consistent with the corresponding prompt.}
\label{tab:sota_compare}
\begin{tabularx}{\linewidth}{lXXXXX}
\toprule
Method 
& \multicolumn{1}{c}{Prompt Align} 
& \multicolumn{2}{c}{Content Align}
& \multicolumn{2}{c}{Style Align} \\
\cmidrule(lr){2-2} \cmidrule(lr){3-4} \cmidrule(lr){5-6}
& CLIP-T($\uparrow$)
& DS($\downarrow$)
& DINO($\uparrow$)
& CLIP($\uparrow$)
& DINO($\uparrow$) \\
\midrule
B-LoRA\cite{frenkel2024implicit}
& 31.6738
& 0.4642
& 0.8367
& \underline{65.7344}
& \underline{0.6777} \\
UnZipLoRA\cite{liu2025unziplora}
& \underline{31.8346}
& 0.4230
& 0.8370
& 61.9083
& 0.6567 \\
StyleID\cite{chung2024styleid}
& 30.5093
& \textbf{0.3179}
& \textbf{0.8849}
& 64.9011
& 0.6634 \\
Dual-LoRA
& 30.5151
& 0.5038
& 0.7947
& 63.3469
& 0.6874 \\
\textbf{AnyStyle (Ours)}
& \textbf{32.5573}
& \underline{0.4210}
& \underline{0.8382}
& \textbf{67.9291}
& \textbf{0.7147} \\
\bottomrule
\end{tabularx}
\vspace{-10pt}
\end{table}

\section{Experiments}

\subsection{Implementation Details}

\textbf{Datasets.} A total of 40 images are collected for the experiments from B-LoRA\cite{frenkel2024implicit}, StyleDrop\cite{sohn2023styledrop}, and EditEval\cite{huang2025diffusion}. These images are evenly divided into two subsets, namely content reference images and style reference images, with 20 images in each subset. Each content reference image is paired with each style reference image, resulting in 400 distinct style transfer combinations.

\noindent
\textbf{Experimental Setup.} Our implementation is based on FLUX\cite{flux2024} model, with the model weights, text encoders, and VAE frozen. During the fine-tuning process, the rank of style LoRA weights is set to 64. We use a learning rate of $5e-5$ and train for 200 steps on a single style reference image. The entire training process takes approximately 3 minutes on a single GPU. For inference, we set the total number of timesteps to 28, with target prompt guidance values of 5. The hyperparameters are set as follows: the maximum and minimum modulation strengths $\beta_{\max}$ and $\beta_{\min}$ are set to 0.7 and 0.3, respectively. All experiments are conducted on an NVIDIA RTX 5880 Ada GPU with 48GB of memory.


\noindent
\textbf{Evaluation Metrics.} Following previous studies\cite{liu2025unziplora,chen2025consislora,yang2025splitflux}, we adopt CLIP\cite{radford2021learning}, PSNR, LPIPS\cite{zhang2018lpips}, MSE, DINO\cite{oquab2024dino} and DS\cite{fu2023dreamsim} as our evaluation metrics. Specifically, the evaluation scores are decomposed into content and style components, and the generated style transfer results are assessed separately against the content reference image and the style reference image, yielding the corresponding content and style scores. In addition, a score is set to evaluate the text consistency, reflecting the semantic matching between prompt and image result.

\begin{wrapfigure}{r}{0.45\textwidth}  
    \vspace{-25pt}   
    \centering
    \includegraphics[width=\linewidth]{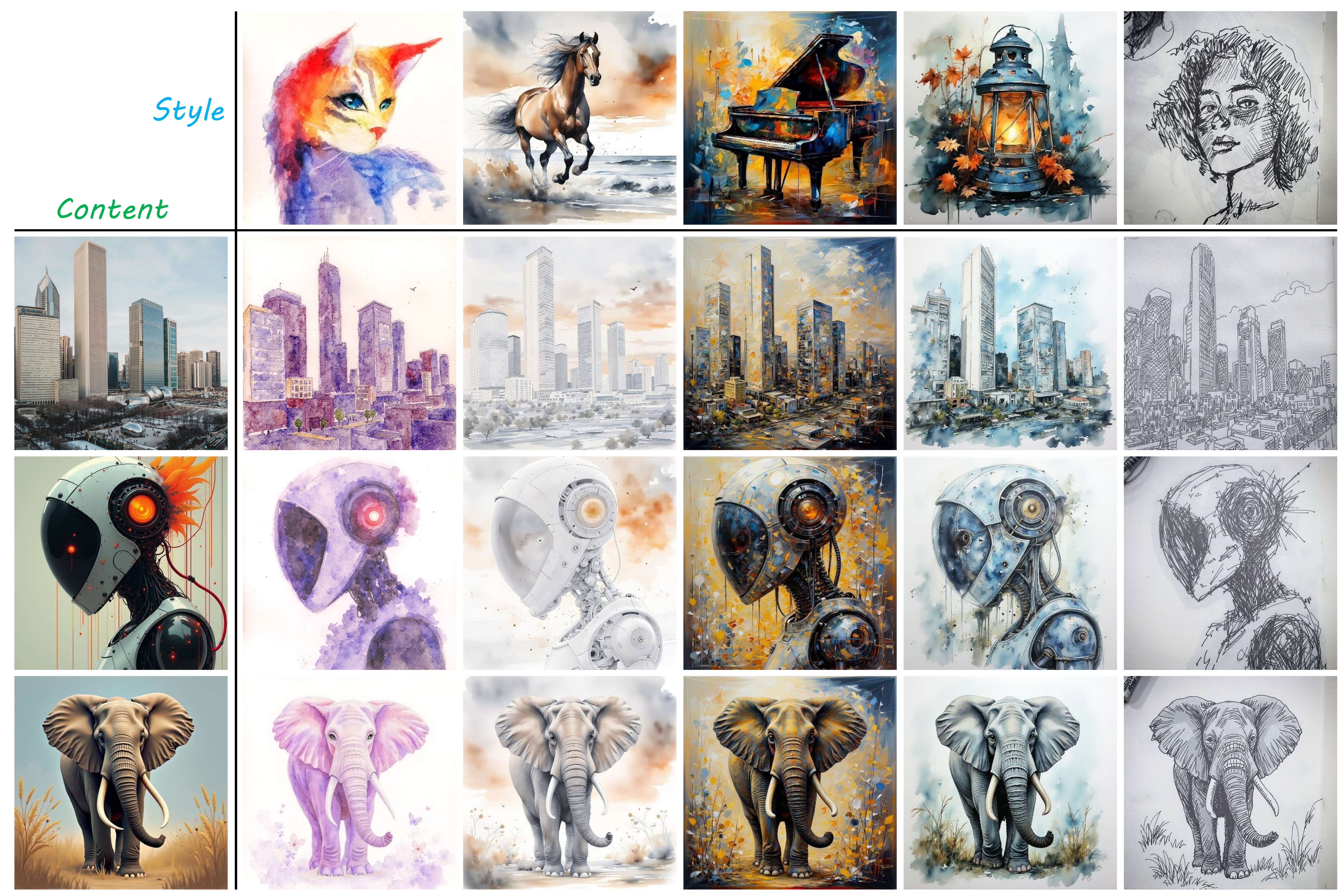} 
    \caption{Additional visual examples of style transfer of our method.}
    \label{fig:more_style_transfer_visualization}
    \vspace{-15pt}
\end{wrapfigure}

\subsection{Results}

\textbf{Comparison with Existing Methods.} We compare AnyStyle with recent state-of-the-art methods for image-guided style transfer. The quantitative results presented in \cref{tab:sota_compare} demonstrate that our approach achieves strong performance in most of the evaluation metrics. Although StyleID\cite{chung2024styleid} attains the highest score in content alignment, it performs worst in prompt alignment and style alignment, which is consistent with the observations reported in ConsisLoRA\cite{chen2025consislora}. Excluding this particular case, our method surpasses all baseline approaches in prompt, style, and content alignment. To validate the effectiveness of the proposed single-LoRA strategy, we conduct a comparative study against a dual-LoRA approach implemented on FLUX. The quantitative results are presented in the fourth method of \cref{tab:sota_compare}, together with the qualitative comparisons shown in \cref{fig:sota_visualization}, which indicate that our method outperforms the "Dual-LoRA" baseline in terms of both content consistency and stylistic coherence. These results validate the effectiveness of the proposed single-LoRA strategy and attention modulation mechanism. The qualitative comparisons shown in \cref{fig:sota_visualization} further support these conclusions, indicating that our approach achieves higher semantic precision in both content and style, resulting in more natural and coherent visual 
outcomes. More visualizations provided in \cref{fig:more_style_transfer_visualization}.

\begin{table}[t]
\centering
\footnotesize
\caption{Quantitative comparison of image generation results with and without loading the style LoRA weights. All other experimental settings are kept identical. The results demonstrate the impact of style-specific adaptation on style alignment.}
\label{tab:ablation_if_lora}
\begin{tabularx}{\linewidth}{lXXXX}
\toprule
Method
& CLIP-T($\uparrow$)
& CLIP-Style($\uparrow$)
& DINO-Style($\uparrow$) \\
\midrule
w/\,\; Style LoRA\quad\quad
& 32.0717\textcolor{green}{$\uparrow$}
& 68.2336\textcolor{green}{$\uparrow$}
& 0.7234\textcolor{green}{$\uparrow$} \\
w/o Style LoRA\quad\quad
& 29.9431
& 57.3314
& 0.6170 \\
\bottomrule
\end{tabularx}
\end{table}

\begin{wrapfigure}{r}{0.45\textwidth}  
    \vspace{-21pt}   
    \centering
    \includegraphics[width=\linewidth]{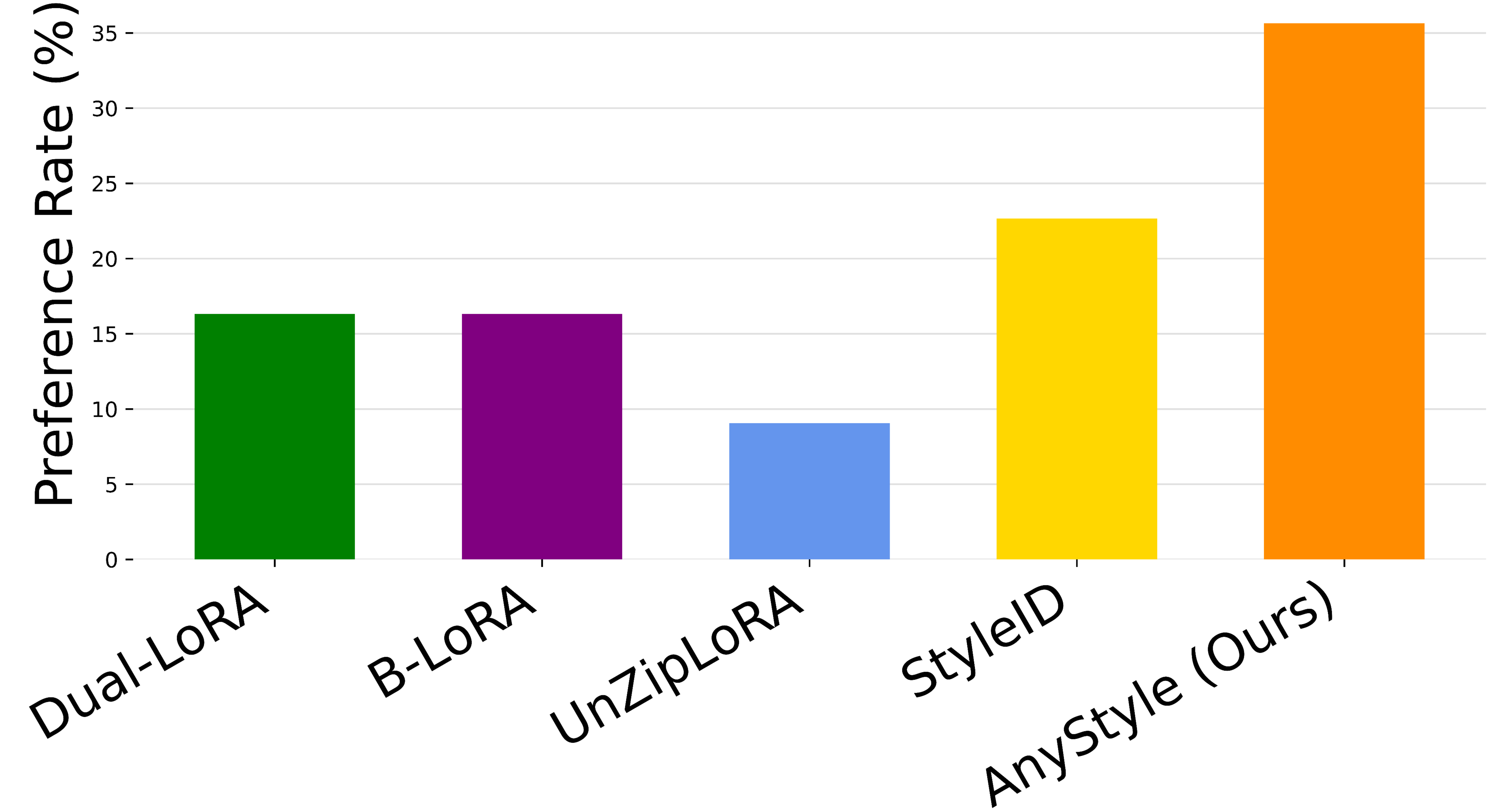} 
    \caption{User study. Our AnyStyle receives the highest number of user selections.}
    \label{fig:userstudy}
    \vspace{-30pt}  
\end{wrapfigure}

\noindent
\textbf{User Study.} To complement quantitative evaluation, we conduct a user study assessing the perceptual quality of AnyStyle against competing methods. A total of 70 participants are presented with the content reference image, style reference image, together with editing results from different methods displayed in randomized order. For each case, participants are asked to select the most visually satisfactory result, with a "Not Sure" option provided to avoid forced choices. Each participant evaluated approximately 10 sets. As shown in \cref{fig:userstudy}, AnyStyle receives the highest preference rate.

\subsection{Ablation Study}

\begin{figure*}[t]
    \centering
    \includegraphics[width=\textwidth]{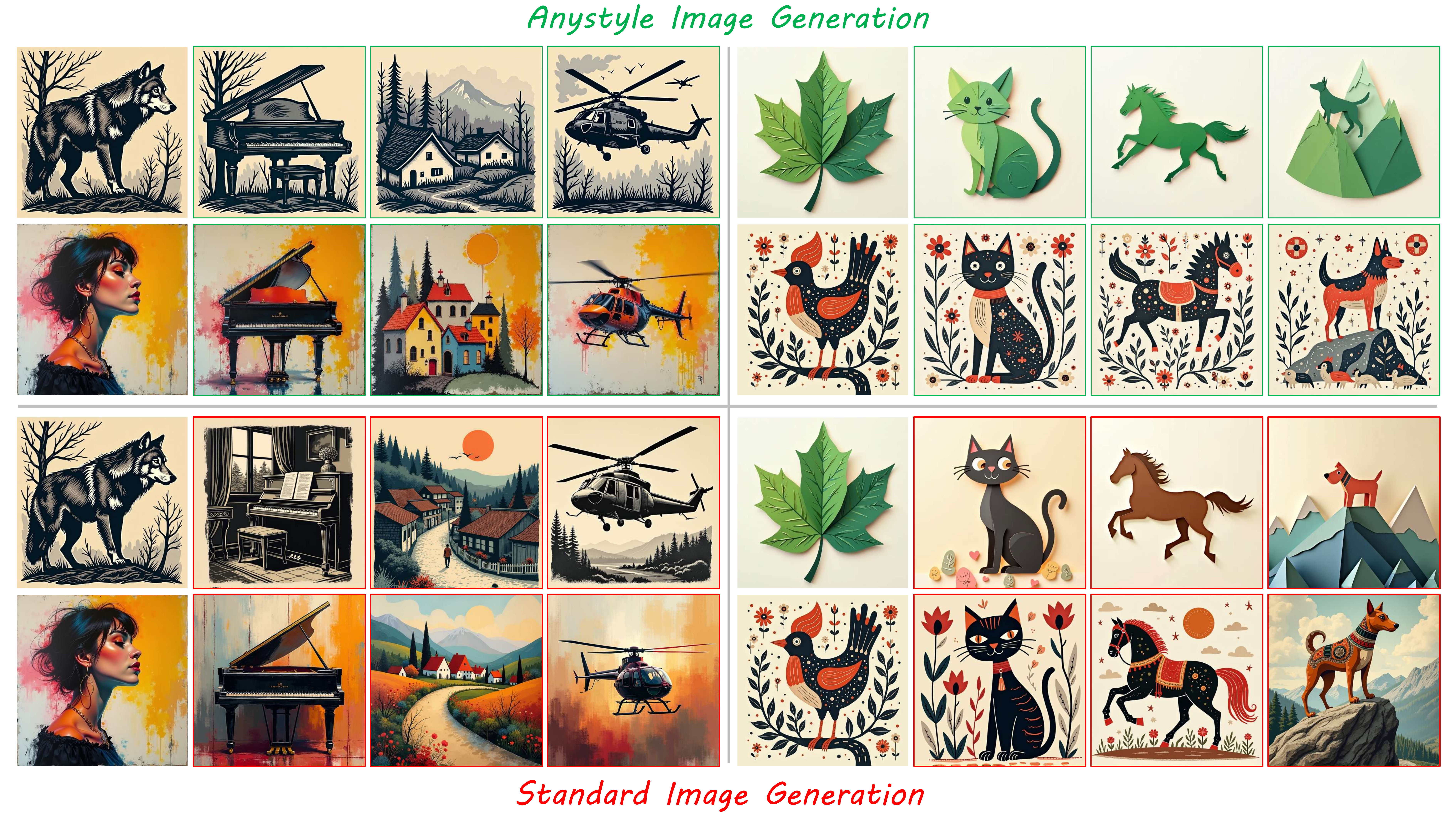}
    \caption{Standard image generation vs. AnyStyle generation with style LoRA weights. The first column of each group is the style reference image. Under the same prompt "A <c> in <s> style", standard sampling without style LoRA (red boxes) yields diverse but stylistically inconsistent results, whereas sampling with the proposed style LoRA (green boxes) produces images with consistent style across different contents.}
    \label{fig:style_alignment_generation}
    \vspace{-5pt}
\end{figure*}

\textbf{Generation of Style Consistency.} To evaluate the effectiveness of the fine-tuned style LoRA weights, we design a controlled comparative experiment. Under an identical textual prompt, images are generated using two configurations: one with the style LoRA weights loaded and one without loading the style LoRA weights. All other model parameters and inference settings are kept fixed to ensure a fair comparison. The generated results are subsequently compared with the corresponding style reference image. As shown in \cref{tab:ablation_if_lora}, the configuration with the style LoRA weights achieves substantially higher stylistic consistency with the style reference image, and a higher text consistency score. This conclusion is further supported by the visual comparisons in \cref{fig:style_alignment_generation}, where the configuration with style LoRA weights loaded exhibits more consistent and stable style.

\begin{figure}[t]
\centering
\includegraphics[width=\linewidth]{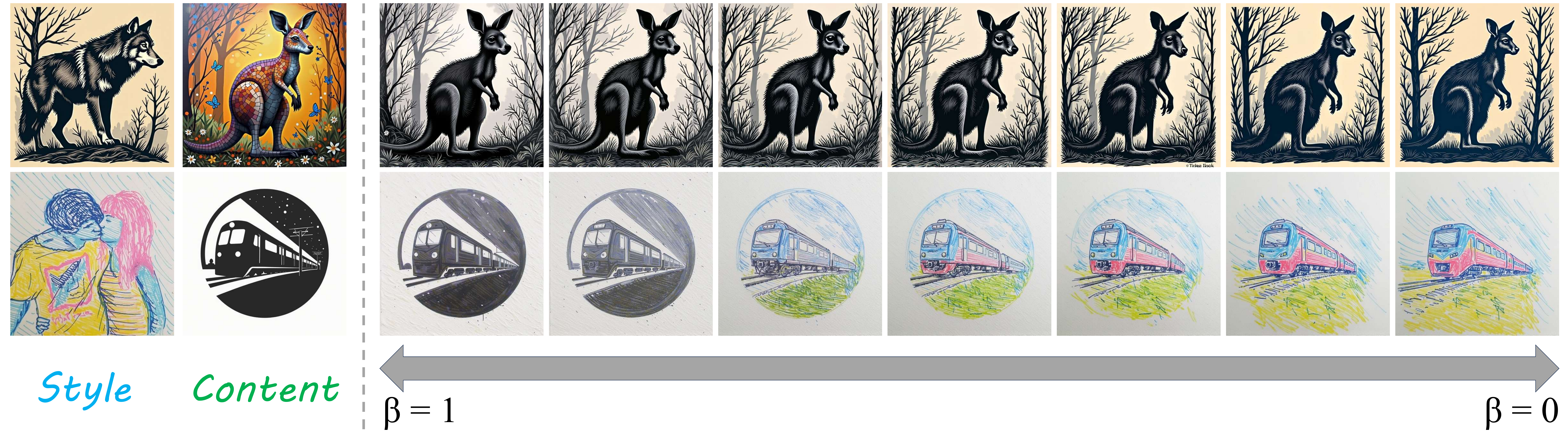}
\caption{Parametric ablation study of the structural guidance strength $\beta_{\max}$ and $\beta_{\min}$.}
\label{fig:beta_ablation}
\end{figure}

\noindent
\textbf{Analysis of Structural Guidance Strength.}
The modulation parameters $\beta_{\min}$ and $\beta_{\max}$ jointly determine the structural control capacity. We conduct an exhaustive parametric sweep to analyze their impact on synthesis characteristics. As visualized in Fig.~\ref{fig:beta_ablation}, systematically decreasing $\beta$ values results in a smooth and continuous transition from heavily content-preserving visual generations to style-dominant outputs. This indicates that AnyStyle offers a highly controllable framework capable of modulating structural preservation boundaries according to diverse user preferences.

\begin{table}[t]
\centering
\footnotesize
\caption{Quantitative comparison under different attention modulation.}
\label{tab:qkv_modulation}
\begin{tabularx}{\linewidth}{lXXXXX}
\toprule
Method 
& \multicolumn{1}{c}{Prompt Align} 
& \multicolumn{2}{c}{Content Align}
& \multicolumn{2}{c}{Style Align} \\
\cmidrule(lr){2-2} \cmidrule(lr){3-4} \cmidrule(lr){5-6}
& CLIP-T($\uparrow$)
& DS($\downarrow$)
& DINO($\uparrow$)
& CLIP($\uparrow$)
& DINO($\uparrow$) \\
\midrule
Query 
& \textbf{32.5573}
& \textbf{0.4210}
& \textbf{0.8382}
& \underline{67.9291}
& \underline{0.7147} \\
Key 
& 32.2734
& 0.5198
& 0.7827
& \textbf{68.4408}
& \textbf{0.7225} \\
Query + Key \quad\quad\quad
& \underline{32.4606}
& \underline{0.4384}
& \underline{0.8333}
& 67.1565
& 0.7034 \\
\bottomrule
\end{tabularx}
\end{table}

\begin{figure*}[t]
    \centering
    \includegraphics[width=0.8\textwidth]{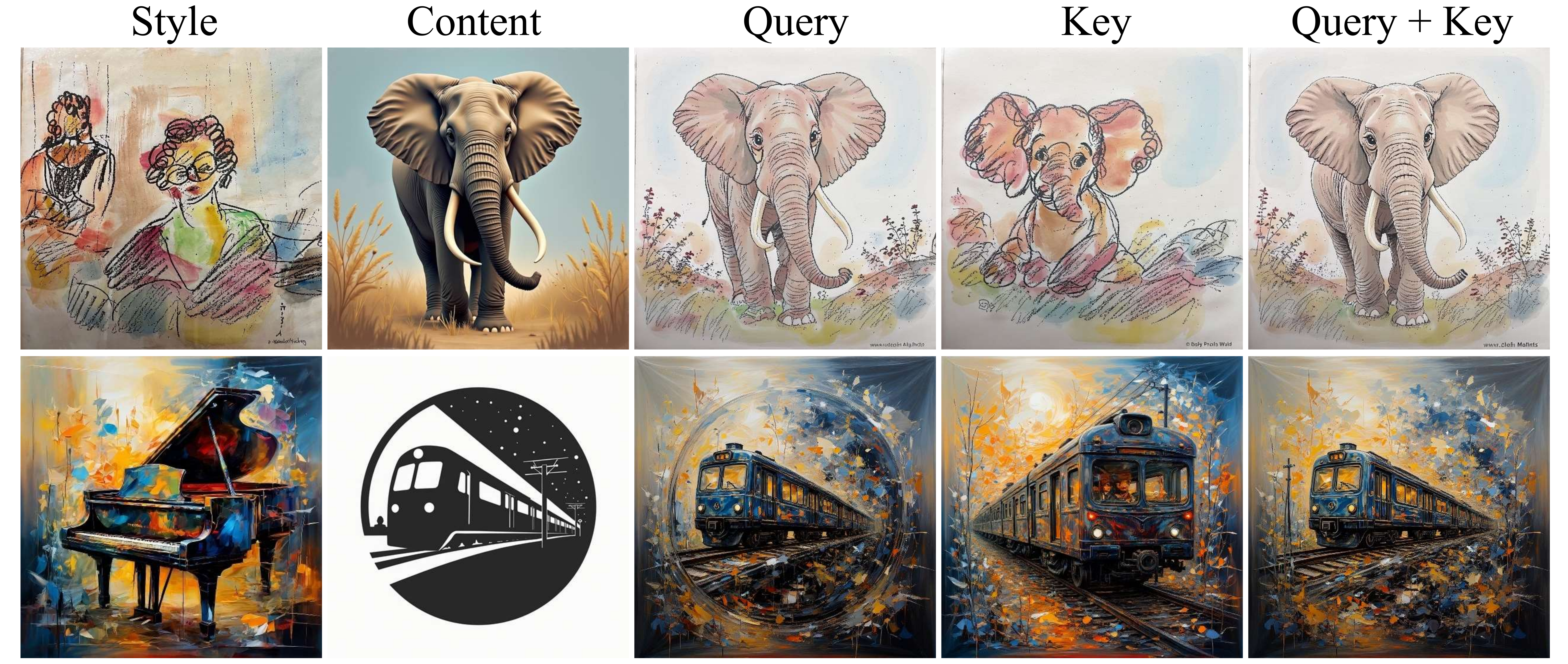}
    \caption{Visualization of style transfer under different attention modulations. Key-only modulation yields strong stylistic consistency but weakens content preservation. Modulating both query and key improves content preservation over the key-only variant, yet reduces stylistic consistency. In contrast, query-only modulation better balances content and style, achieving the most effective overall style transfer performance.}
    \label{fig:qk_ablation}
\end{figure*}

\noindent
\textbf{Effect of Attention Modulation.}
To investigate the influence of different attention modulation strategies on the final performance, we conduct a series of controlled experiments. Based on the single-LoRA framework, various attention modulation schemes are introduced during the denoising process. As reported in \cref{tab:qkv_modulation}, applying modulation solely to the key tensors yields relatively high style alignment scores but leads to substantially degraded content preservation. When both query and key tensors are modulated, content-related metrics improve compared to key-only modulation, yet the gains in style alignment remain limited. In contrast, modulating only the query tensors achieves the best overall performance across evaluation metrics. This outcome is consistent with our expectations. Query-only modulation maximally exploits structural guidance, effectively preserving content structure. At the same time, it restricts the propagation of extraneous information from the content reference image into the main branch, thereby preventing unintended stylistic leakage and resulting in superior style transfer performance. Similarly, as illustrated in \cref{fig:qk_ablation}, query-only modulation demonstrates the most favorable trade-off between content preservation and style alignment.

\begin{figure}[t]
\centering
\includegraphics[width=0.8\linewidth]{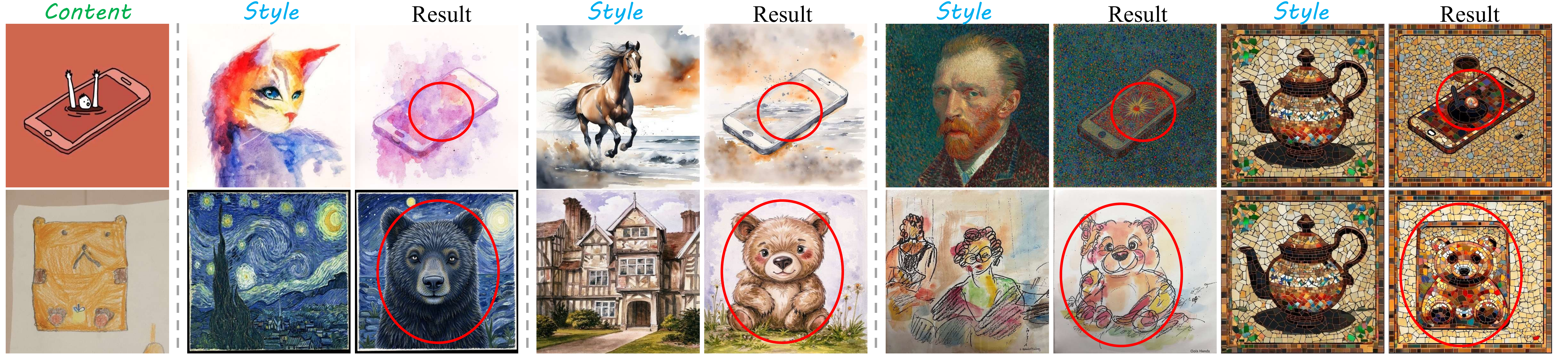}
\caption{Typical failure cases visualization of our proposed method.}
\label{fig:failure_case}
\end{figure}

\noindent
\textbf{Failure Cases.}
Typical failure cases tend to manifest when the pre-trained model fails to reliably recognize highly complex and non-standard semantic structures. For instance, when encountering uncommon semantics rarely seen in the training data, such as "a cartoon picture depicting a person submerged in a mobile phone screen" and "a colored pencil drawing depicting a cute little bear", AnyStyle exhibits an inability to comprehend such intricate textual descriptions. An illustrative example of this behavior is explicitly detailed in Fig.~\ref{fig:failure_case}.

\section{Conclusion}
In this paper, we propose AnyStyle, a streamlined and effective framework for image-guided style transfer with improved controllability and stability. By employing a single-adapter strategy for consistent style generation, we avoid the complexities of multi-adapter fusion and the resulting entanglement between content and style. Furthermore, AnyStyle integrates attention modulation to provide precise structural guidance from the content reference image, ensuring faithful preservation of semantic layout and structure. Experiments demonstrate that our method achieves competitive performance compared to existing approaches, with enhanced coordination among prompt adherence, style consistency, and content fidelity in both quantitative metrics and perceptual evaluations. 

\footnotesize 
\section{Acknowledgments}
This work was supported by Guangdong Basic and Applied Basic Research Foundation (Nos. 2024A1515140109 and 2023A1515110695).

\clearpage

\bibliographystyle{splncs04}
\bibliography{main}

\clearpage
\appendix

\title{AnyStyle: A Single LoRA is Sufficient for Image-Guided Style Transfer (Supplementary Material)} 

\titlerunning{AnyStyle}

\author{Yongwen Lai \orcidlink{0009-0007-2597-5065} \and
Chaoqun Wang\thanks{Corresponding author.} \orcidlink{0000-0002-4649-5518}}

\authorrunning{Yongwen Lai, Chaoqun Wang}

\institute{School of Artificial Intelligence, South China Normal University, Guangzhou, China 
}

\maketitle

\appendix

\section{Algorithmic Details}
\label{sec:algorithm_detail}
To provide a more comprehensive view of the core pipeline of AnyStyle, we present the full algorithm of AnyStyle in Algorithm~\ref{alg:anystyle}. Given the structural guidance latent $Z^{c}$, style target $Z^{s}$, total diffusion timesteps $T$, and the dynamic modulation thresholds $\beta_{max}$ and $\beta_{min}$, our method modulates the latent representations to ensure semantic fidelity and style dominance. First, the task-specific style weights $\Delta W$ are extracted using DreamBooth optimization on the style target:
\begin{equation}
    \Delta W \leftarrow \text{DreamBooth}(Z^{s}), \quad \Theta \leftarrow \theta + \Delta W,
\end{equation}
where $\theta$ denotes the pre-trained model parameters. At each sampling step $t$, the structural modulation coefficient $\beta$ linearly transitions according to the current stage of synthesis. The content queries $Q_c$ are explicitly extracted via an auxiliary denoising path, which subsequently guides the update of the target latent representation $Z_t$. The complete sequence of operations and the structured routing flow are explicitly detailed in the algorithm below.

\begin{algorithm}[h]
\centering
\footnotesize
\raggedright
\noindent\textbf{Input:} $Z^{c}, Z^{s}, \{t_i\}_{i=0}^T, n,\beta_{max}, \beta_{min}$ \\
\noindent\textbf{Output:} Style Transfer Result $Z_0$ \\
$\Delta W \leftarrow \text{DreamBooth}(Z^{s})$;\;$\Theta \leftarrow \theta + \Delta W$\; \\
Initialize $Z_T \sim \mathcal{N}(0,I)$; $ \epsilon \sim \mathcal{N}(0,I)$ \\
\textbf{for} $i=n$ to $1$: \\
\quad $\beta \leftarrow \beta_{max} \cdot ({t_i}/T) + \beta_{min} \cdot (1 - {t_i}/T)$ \\
\quad $Z^{c}_{t_i} \leftarrow (1-{t_i}) \cdot Z^{c} + {t_i} \cdot \epsilon$ \\
\quad $V^{c}_{t_i} \leftarrow V_{\Theta}(Z^{c}_{t_i},{t_i}) \implies \text{extract } Q_{\text{c}}$; \quad $V_{t_i} \leftarrow V_{\Theta}(Z_{t_i},{t_i},Q_{\text{c}})$ \\
\quad $Z_{t_{i-1}} \leftarrow Z_{t_i} - \Delta {t} \cdot V_{t_i}$ \\
\textbf{return} $Z_0$ \\
\caption{Full AnyStyle Algorithm}
\label{alg:anystyle}
\end{algorithm}

\begin{table}[t]
\centering
\caption{Quantitative benchmarking against alternative style LoRA frameworks.} 
\label{tab:style_lora}
\begin{tabular}{l|ccc}
\toprule
Method & CLIP-T$\uparrow$ & CLIP-S$\uparrow$ & DINO-S$\uparrow$ \\ 
\midrule
w/o LoRA & 29.9431 & 57.3314 & 0.6170 \\ 
B-LoRA   & 31.7421 & \underline{68.1875} & 0.7070 \\
UnZipLoRA& \underline{32.0033} & 67.0723 & \underline{0.7153} \\ 
\hline 
Ours     & \textbf{32.0717} & \textbf{68.2336} & \textbf{0.7234} \\ 
\bottomrule
\end{tabular}
\end{table}

\begin{figure}[t]
\centering
\includegraphics[width=\linewidth]{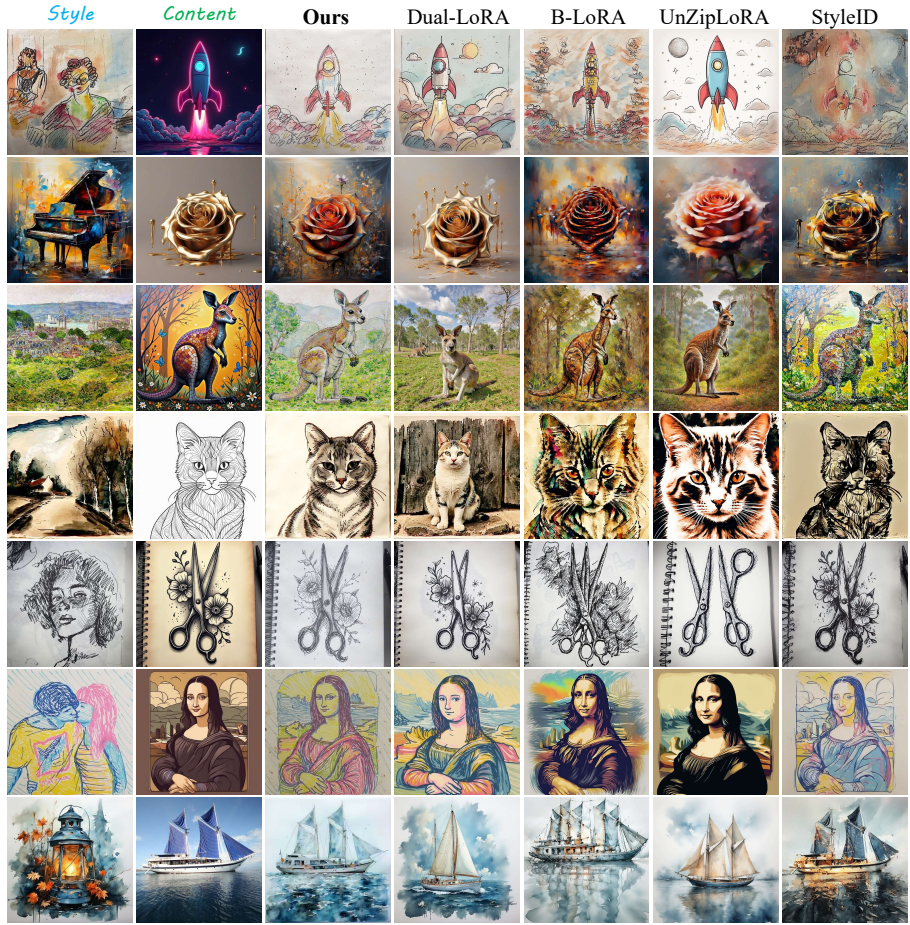}
\caption{Qualitative comparisons against contemporary and classical baseline frameworks.}
\label{fig:sota_vis}
\end{figure}

\section{Comparison of Style LoRA Representation Capabilities} 
To evaluate the effectiveness of our style feature learning, we compare the LoRA weights generated by AnyStyle against contemporary state-of-the-art style-specific low-rank adaptation methods, specifically B-LoRA\cite{frenkel2024implicit} and UnZipLoRA\cite{liu2025unziplora}. As reported in Table~\ref{tab:style_lora}, AnyStyle consistently outperforms these alternative representations across all metrics. This significant margin demonstrates that our optimization captures artistic textures and stylistic nuances more robustly, yielding superior style dominance without sacrificing text alignment.

\section{Generalizability Across Diffusion Backbones} 
To demonstrate that AnyStyle is not restricted to DiT-based architectures (e.g., Flux\cite{flux2024}), we deploy our framework onto a U-Net backbone utilizing SDXL\cite{podell2024sdxl}. As summarized in Table~\ref{tab:sdxl_generalization}, AnyStyle achieves highly competitive metrics across all evaluations, establishing that our core attention modulation mechanism generalizes seamlessly across disparate generation architectures.

\begin{table}[t]
\centering
\caption{Quantitative performance evaluations on alternative diffusion baselines and classical style transfer paradigms.}
\label{tab:sdxl_generalization}
\begin{tabular}{l|ccccc}
\toprule
Method & CLIP-T$\uparrow$  & DS-C$\downarrow$ & DINO-C$\uparrow$ & CLIP-S$\uparrow$&  DINO-S$\uparrow$ \\
\midrule
CAST\cite{zhang2022domain}  & 29.6117& \underline{0.3259}& \underline{0.8738} & 60.8029  & 0.6467 \\
S2WAT\cite{zhang2024s2wat} & 29.8412& \textbf{0.3142}& \textbf{0.8874} & 58.2346  & 0.6367  \\
\hline
AnyStyle(SDXL) & \underline{32.2765}& 0.4109  & 0.8341& \underline{66.4223} & \underline{0.6936}  \\
AnyStyle(FLUX) & \textbf{32.5573}& 0.4210& 0.8382  & \textbf{67.9291} & \textbf{0.7147}  \\
\bottomrule
\end{tabular}
\end{table}

\section{Comparison against Classical Style Transfer Paradigms} 
We compare our approach against prominent classical methods including CAST\cite{zhang2022domain} and S2WAT\cite{zhang2024s2wat}. As detailed in Table~\ref{tab:sdxl_generalization} and visual results in Fig.~\ref{fig:sota_vis}, while classical optimization schemes maintain higher local content preservation indices (e.g., lower DS-C and higher DINO-C), they underperform significantly regarding text prompt alignment (CLIP-T) and complex artistic stylization (CLIP-S). AnyStyle achieves a much more optimized balance between structural preservation and expressive texture migration.

\section{Computational Efficiency Profile} 
We benchmark the computational overhead and inference times of AnyStyle against representative baselines. The performance evaluations are carried out on an identical hardware configuration and reported in Table~\ref{tab:computational_time}.

Although training-free CNN/Transformer pipelines naturally exhibit lower latencies, they fail to synthesize complex, high-fidelity fine-grained patterns. Compared to alternative fine-tuning methods, AnyStyle removes the requirement to optimize a separate content-specific LoRA weight for each test case, thus significantly mitigating resource expenditures while offering a highly efficient runtime footprint.

\begin{table}[h]
\centering
\caption{Inference execution time comparison per image execution.}
\label{tab:computational_time}
\begin{tabular}{lc} 
\toprule
Method & Inference Time (s/Image) \\
\midrule
CAST\cite{zhang2022domain} / S2WAT\cite{zhang2024s2wat} / StyleID\cite{chung2024styleid}  & 4 / 5 / 12  $\pm$ 1 \\
Dual-LoRA / B-LoRA\cite{frenkel2024implicit} / UnZipLoRA\cite{liu2025unziplora} & 11 / 13 / 25 $\pm$ 1 \\
\midrule
\textbf{AnyStyle (Ours)} & 18 $\pm$ 1 \\
\bottomrule
\end{tabular} 
\end{table}



\begin{figure}[t]
  \centering
  \includegraphics[width=\textwidth]{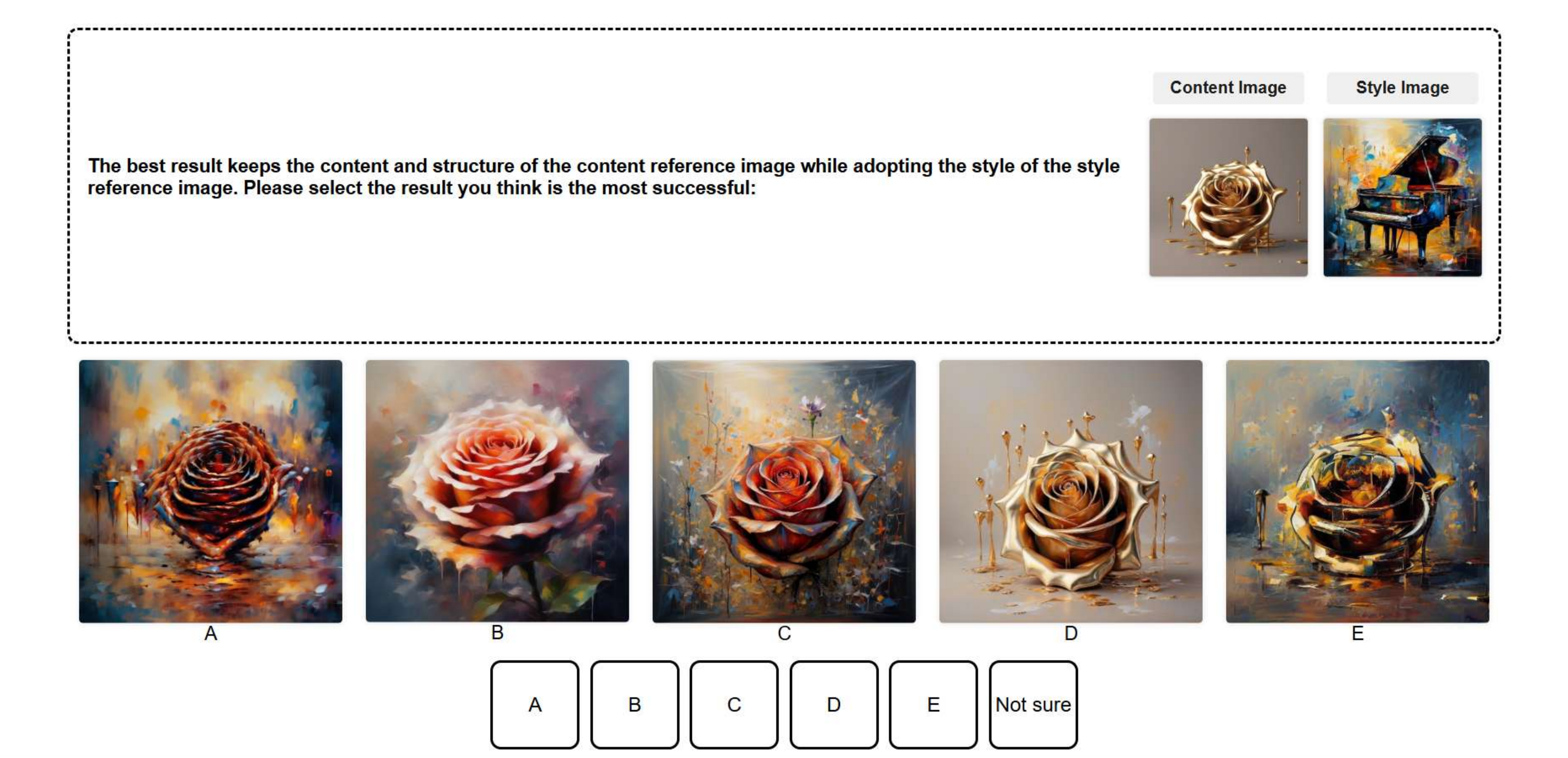}
  \caption{User study interface. Participants were shown the content image, style image, and outputs from various methods, and were asked to select the most successful result.} 
  \label{fig:userstudy_supplement}
  \vspace{-10pt}
\end{figure}

\section{User Study Interface} 
The user study interface is shown in Fig.~\ref{fig:userstudy_supplement}.
Each participant was given clear instructions, and in each trial, was presented with the content image, style image and outputs from multiple methods (including ours), displayed in randomized order to reduce positional bias.
Participants were asked to select the result that best reflected the target prompt while preserving the essential content of the source image.
A ``Not Sure'' option was provided to accommodate ambiguous or low-confidence cases.
Invalid entries, \emph{i.e.}, participants selecting the same option across all trials, were excluded from the final analysis.


\end{document}